\documentclass{article} 
\usepackage{iclr2016_conference,times}
\usepackage{hyperref}
\usepackage{enumitem}
\usepackage{amsmath,amssymb,amsthm}
\usepackage{url}
\usepackage{color}
\usepackage{verbatim}
\usepackage{graphicx}
\usepackage{subfigure}
\usepackage{caption}

\title{A Distribution Adaptive Framework for Prediction Interval Estimation Using Nominal Variables}

\author{
	Ameen Eetemadi \\
	Department of Computer Science\\
	University of California\\
	Davis, CA 95616 \\
	\texttt{eetemadi@ucdavis.edu} \\
	\And
	Ilias Tagkopoulos \\
	Department of Computer Science \\
	University of California \\
	Davis, CA 95616 \\
	\texttt{itagkopoulos@ucdavis.edu} \\
}

\begin{document}

	\maketitle
	
	\begin{abstract}
Proposed methods for prediction interval estimation so far focus on cases where input variables are numerical. In datasets with solely nominal input variables, we observe records with the exact same input $x^u$, but different real valued outputs due to the inherent noise in the system. Existing prediction interval estimation methods do not use representations that can accurately model such inherent noise in the case of nominal inputs. We propose a new prediction interval estimation method tailored for this type of data, which is prevalent in biology and medicine. We call this method Distribution Adaptive Prediction Interval Estimation given Nominal inputs (DAPIEN) and has four main phases. First, we select a distribution function that can best represent the inherent noise of the system for all unique inputs. Then we infer the parameters $\theta_i$ (e.g. $\theta_i=[mean_i, variance_i]$) of the selected distribution function for all unique input vectors $x^u_i$ and generate a new corresponding training set using pairs of $x^u_i, \theta_i$. III). Then, we train a model to predict $\theta$ given a new $x_u$. Finally, we calculate the prediction interval for a new sample using the inverse of the cumulative distribution function once the parameters $\theta$ is predicted by the trained model. We compared DAPIEN to the commonly used Bootstrap method on three synthetic datasets. Our results show that DAPIEN provides tighter prediction intervals while preserving the requested coverage when compared to Bootstrap. This work can facilitate broader usage of regression methods in medicine and biology where it is necessary to provide tight prediction intervals while preserving coverage when input variables are nominal.

	\end{abstract}
	
	\section{Introduction}
	Although most of the well-known regression methods, are designed to provide point prediction for a real valued variable, this is not always sufficient. There are many cases where the stakeholders are interested in knowing the precision of individual predictions. In the machine learning literature this is sometimes referred to as Conformal Prediction (CP) \cite{papadopoulos2011reliable} and others simply refer to it as providing Prediction Interval (PI) \cite{khosravi2011comprehensive}. In the general single-task learning (STL) problem, a regression CP method trains a predictive model. The predictive model takes an input feature vector $x \in \mathbb{R}^d$, a desired confidence level $\alpha \in [0,100]$ and provides a PI $[y_l, y_u]$ for the output variable $y \in \mathbb{R}$. It is conceivable to have a model which would predict multiple intervals, but we only study the case which providing a single interval is sufficient. Compared to point prediction, such precise information would be much more useful for human decision making in various applications such as clinical diagnosis, financial services and experimental design. PIs also enable getting meaningful predictions by stacking layers of - independently trained - predictors on top of each other. \\
	In this work, we provide a novel method called DAPIEN tailored to datasets with nominal input variables. Although  DAPIEN can incorporate various regression techniques, we only use Artificial Neural Networks (ANN) in our experiments. This is in part due to the fact that, from a theoretical standpoint, ANNs can model any non-linear function and in the last ten years, there has been record breaking achievements through the application of ANNs particularly in speech recognition, computer vision, machine translation and predictive genomics in the context of deep learning \cite{hinton2006fast, seide2011conversational, lee2009convolutional, alipanahi2015predicting}. \\
	
	Section \ref{relwork} contains a discussion on existing methods for PI prediction with focus on ANNs. The detailed description DAPIEN is provided in section \ref{method}. Experimental results using synthetic data is provided in section \ref{result}. Conclusion and future work is briefly discussed in section \ref{conclusion}.
	\section{Related Work}\label{relwork}
	An extensive review for neural network-based prediction intervals is provided in \cite{khosravi2011comprehensive}. Next we provide a concise summary of PI assessment measures and also discuss the latest developments in PI prediction methods, some of which have not been reviewed together before.
	
	\subsection{PI Assessment Measures}
	When it comes to assessing the quality of PIs, it is common to solely focus on the overall coverage of the calculated intervals [ToDo: Add ref from \cite{khosravi2011comprehensive}]. Such an approach can lead into favoring methods that provide wide intervals and hence provide minimal specificity for the predicted outcome. The following are among the most informative PI quality measures.
	\begin{itemize}
		\item{Prediction Interval Coverage Probability(PICP)}, is calculated by measuring the fraction of target values covered by the PIs
		\begin{equation}
			PICP = \frac{\left\vert{\{y_i| y_i \in Targets, y_i \in [\hat{y_{il}}, \hat{y_{iu}}]}\right\vert}
			{\left\vert{Targets}\right\vert}
		\end{equation}
		
		\item Mean PI Width(MPIW), quantifies the overall width of the PIs
		\begin{equation}
			MPIW = \frac{\sum_{i=1}^{N_{Targets}} (\hat{y_{iu}} - \hat{y_{il}})}
			{\left\vert{N_{Targets}}\right\vert}
		\end{equation}
		
		\item{Normalized MPIW (NMPIW)}, enables comparison of PIs using different datasets
		\begin{equation}
			NMPIW = \frac{MPIW}{Range(y)}
		\end{equation}
		
		\item{Coverage Width-based Criterion (CWC)} combines PICP and NMPIW and serves as a single quality measure to enable draw comparisons for multiple methods
		\begin{equation}
			CWC = NMPIW(1 + \gamma(PICP) e^{-\eta(PICP-\mu)})
		\end{equation}
		\begin{equation}
			\gamma(PICP) = 
			\begin{cases}
				0,& PICP \ge \mu \\
				1,& otherwize
			\end{cases}
		\end{equation}
		A smaller CWC is preferred over larger ones. The $\mu$ parameter, gives the user control over the acceptable PICP. By tuning the $\eta$ parameter, user would be able to define the relative importance of PICP in the overall CWC measure when PICP is bellow the acceptable threshold.
	\end{itemize}
	In this study, we only report the $PICP$ and $MPIW$ measures as they are easier to interpret and describe well what we aim to achieve. Other PI assessment measures can be useful particularly in studies where either large number of various datasets are involved or the PI prediction method involves a feedback loop to use a PI measure such as $CWC$ to tune the prediction process.
	\subsection{Neural Network-based Methods for providing Prediction Intervals} \label{eixisting_PI_methods}
	Before introducing the new DAPIEN method, it is important to understand existing methods to see why a method tailored for categorical dataset is necessary.
	\subsubsection{Bootstrap}
	The most common technique for construction of PIs is the Bootstrap method \cite{heskes1997practical}. It consists of two phases: first, estimate the portion of error which comes from inaccurate modeling and then estimate the error which comes from noise in the observed output itself. In the first phase, the Bootstrap resampling is used to generate $B$ datasets, which are then used to train the prediction method. Although what we describe is applicable in many regressors and classifiers, for the rest of the paper we will focus on artificial neural networks (ANN).

	When provided with a new sample for prediction, all trained ANNs will be used to provide with $B$ predictions. The variance of the predictions represents the modeling error. In the second phase, a separate ANN is trained for predicting the output error. This is done by generating a new data-set in which each sample consists of input features and the residual errors. For each sample, the residual error is generated by subtracting the modeling error from the prediction error in the first phase. Given a new sample, the overall error is estimated by adding errors from both phases. The following PI would is provided:
	\begin{equation}
		\mu - c_{confidence} \sigma \le \hat{y} \le \mu + c_{confidence} \sigma
	\end{equation}
	The $C_{confidence}$ value is calculated using the t-table, the provided confidence level and $B$ degrees of freedom. The $\mu$ is the mean of the predicted value as calculated in the first phase. The value of $\sigma$ is the estimated by adding the estimated error from both phases.\\
	
	\subsubsection{Bayesian and Delta}
	Although the Bayesian \cite{bishop1995neural} and Delta \cite{hwang1997prediction} are very different from each other, they have similar properties. Similar to the Bootstrap, they both distinguish between the modeling error and data error.  But in contrast to Bootstrap they assume normal distribution for the error (with Delta having stronger assumptions), regardless of the output values. They also include the costly calculation of the Hessian matrix. They both provide high quality PIs when the assumptions for distribution of error are valid. The Bayesian method for training ANN, has particularly strong mathematical foundation and good generalization error \cite{khosravi2011comprehensive}.
	
	\subsubsection{MVE}
	In Mean and Variance Estimation method \cite{nix1994estimating}, it is also assumed that the errors are normally distributed around the mean of the target. Although in the Bayesian and Delta methods assume fixed variance, MVE assumes that the variance is also a function of the input. Therefore, it trains two distinct neural networks for estimating the mean and variance of the output. One drawback for this method, is ignoring the modeling error \cite{khosravi2011comprehensive} which affects the quality of the predicted intervals.
	
	\subsubsection{Back-propagation of Pseudo-Errors}
	In \cite{ding2003backpropagation}, the authors consider a case in which the distribution of error is not necessarily normal and can be skewed. The idea is to integrate the box-cox transformation \cite{sakia1992box} during training to model non-Gaussian distribution of noise. It will then use the Bootstrap method for providing PIs.
	
	\subsubsection{Combined PIs}
	\cite{khosravi2011comprehensive} proposes an ensemble method using Bootstrap, Bayesian, Delta and MVE methods. First the dataset is separated to two parts $D_1, D_2$. $D_1$ is used to train all methods. Then an ensemble approach is taken by providing new PI which is a linear combination of the intervals provided by all four approaches. The parameters of the linear combination is estimated in an optimization procedure to minimize the overall $CWC$ measure over $D_2$. Due to the form of $CWC$ function, gradient descent cannot be used and instead a Genetic Algorithm approach is taken. Authors show that the Combined PI approach consistently outperforms the individual methods when compares based on the $CWC$ measure.
		
	\section{Method} \label{method}
	All PI prediction methods described in \ref{eixisting_PI_methods}, deal with the case where the input vector is real valued. Hence they lack the ability to measure and model the variance of output within a group of samples with identical nominal inputs. In other words, although all nominal input vectors can be represented with real valued vectors, but by throwing away the information within each group of samples (e.g. mean, variance, skewness of the output variable) the quality of prediction intervals can be undermined.\\
	Here we discuss the problem of estimating confidence interval for an output variable $y \in \mathbb{R}$, given the input binary vector $ \mathbf{x} = (x_1, x_2, ..., x_d)$ where $x_i \in \{0, 1\}$ representing nominal features. Assume that the output variable $y \in \mathbb{R}$, is a single scalar value. However, the same method can be generalized to multi-task learning for predicting multiple output variables. Also assume that the distribution of $y$, follows the same form of distribution function regardless of the input parameters, however the parameters of the distribution function can vary depending on the input vector. We show in section \ref{mpractical}, one way to use this method when distributions do not follow the same function.
	
	\subsection{General procedure} \label{moutline}
	The following describes the method outline:\\
	\begin{enumerate}
		\item Given a distribution function $f_{\theta}(y)$, where $\int_{-\infty}^{+\infty} f_{\theta}(y) dy = 1$ and the training set $D^t = \{(x_i, y_i)\}, i=1 \dots N^{total}$:
		\begin{enumerate}
			\item Identify the unique input training vectors $x^{u}, u=1 \dots p$. For each $x^u$, there is a corresponding $y^u$ which consists of all $y_i$ records which share the same $x_i$
			\item Construct p datasets, each containing samples that have the same input vectors, represented by $D^{u} = \{x^u, y^u\}, u=1 \dots p$, such that $x^u \in \mathbb{R}^d, y^u \in \mathbb{R}^{N^u}$ where $ N^u$ is the number of samples with input vector $x = x^u$
			\item \label{mo1c} Using each dataset $D^u$, estimate $\theta^u$, representing the parameters of distributions $f_{\theta^1}(y|x^1) \dots f_{\theta^p}(y|x^p)$. Note that $\theta^u \in \mathbb{R}^k$, where $k$ is the number of parameters in the distribution function $f$
			\item Construct new training set $D_{dist}=\{(x^u, \theta^u)\}, u=1 \dots p$
		\end{enumerate}
		\item \label{mo2} Using the new data set $D_{dist}$, train a model for predicting $\theta$
		\item \label{mo3} Given a new input vector $x$, the trained model in step \ref{mo2}, provides $\hat{\theta}$ representing distribution $f_{\hat{\theta}}(y|x)$
		\item \label{mo4} Given a confidence level $\alpha$ and $f_{\hat{\theta}}(y|x)$, a prediction interval will be calculated.
	\end{enumerate}
	
	Next we describe this method for Gaussian and Gamma distributions. The same procedure can be extended to other types of distribution functions.
	\subsection{Gaussian Distribution Procedure} \label{mpractical}
	Consider the case where a single output variable, follows a Gaussian distribution while the parameters of the distribution may vary given input vector x.
	For providing prediction intervals, the general procedure as in section \ref{mpractical},  will be followed. 
	In step \ref{mo1c}, for each unique $x^u$ the Gaussian distribution parameters $\theta^u = (\mu^u,\ {\sigma^2}^u)$ would be calculated from each dataset $D^u$. After constructing $D_{dist}$ using the estimated values of $\theta$, a predictor would be trained to predict $\theta$ given a new input. 
	\\Finally (as in step \ref{mo4}) in order to provide the prediction interval, the Student's t-distribution will be used. Given a confidence level $\alpha$ and degrees of freedom ($nDF$), the $c_{confidence}$ will be calculated from the t-table. The $nDF$ will be the average number of samples in $D^u$s. Hence the following PI will be provided using the estimated values of $\mu$ and $\sigma$ and the $c_{confidence}$ taken from Student's t-distribution:
	\begin{equation}
		\mu - c_{confidence} \sigma \le y_t \le \mu + c_{confidence} \sigma
	\end{equation}

	[missing citation below]

	\subsection{Gamma Distribution Procedure} \label{mpracticalGamma}
	Although, Gaussian distributions are widely used, they are not applicable to many datasets. For example the steady-state probability distribution of protein number per cell is known to follow a gamma distribution \cite{cai2006stochastic}. For a Gamma probability distribution, the general procedure in section \ref{mpractical},  will be followed. In step \ref{mo1c}, for each unique $x^u$ the Gamma distribution parameters  $\theta^u = (\alpha^u,\ {\beta}^u, {\mu}^u)$ will be estimated ($\alpha$ is the shape parameter, $\beta$ is the scale parameter, and $\mu$ is the location parameter). Then similar to section \ref{mpractical}, $D_{dist}$ will be constructed using the $x^u$ and estimated values of $\theta$. Then a model will be trained using $D_{dist}$ as described in \ref{mpractical}.
	For providing the prediction interval (as in step \ref{mo4}), the inverse gamma distribution function $F{-1}$ would be used (see equation \eqref{eq:gammaInverseEq}). That is, given a new $x$, first  $\hat{\theta} = (\hat{\alpha},\hat{\beta}, \hat{\mu})$, will be predicted. Then using the requested confidence level $c_{confidence}$ (e.g. 0.95), $(\hat{\alpha},\ \hat{\beta},\  \hat{\mu})$ and $F^{-1}$ the prediction interval will be provided as in equation \eqref{eq:gammaPIEq}.
	
	\begin{equation}\label{eq:gammaInverseEq}
		\begin{split}
			& pdf(t ; \alpha,\ \beta,\ \mu) =  \frac{\beta^\alpha}{\Gamma(\alpha)} (t-\mu) e^{-\beta (t-\mu)}\\
			& F^{-1}(p, \alpha,\ \beta,\ \mu) = z \ \ \text{s.t.} \ \ \int_{0}^{z} pdf(t ; \alpha,\ \beta,\ \mu)\ dt = p\\
		\end{split}
	\end{equation}
	\begin{equation}\label{eq:gammaPIEq}
		F^{-1}(\frac{1-c_{confidence}}{2}, \hat{\alpha},\ \hat{\beta},\  \hat{\mu}) \le y_t \le F^{-1}(\frac{1+c_{confidence}}{2}, \hat{\alpha},\ \hat{\beta},\  \hat{\mu})
	\end{equation}
	
	\section{Results}\label{result}
	As an early step in methodical assessment of the performance, synthetic data has been used. This allows us to (i) introduce controlled error to the dataset, (ii) ensure, there is an underlying pattern in the data-set to be discovered by training ANN and (iii) examine and debug a given method in a smaller scale before scaling up.

	Therefore two datasets with distinct characteristics are generated to evaluate the performance of the proposed method against the common Bootstrap method in the case where input variable solely consists of categorical values. Table \ref{tab:PI_SumOfBits_2_all} shows the overall results where the proposed DAPIEN method provides higher PICP levels while maintaining an appropriate MPIW level. The following sections provide more detail about the data generation procedure as well as the figurative illustration of the performance for of each method.
	
	\begin{table}[!htb]
		\begin{center}
			
			\begin{tabular}{| l | l| l| l | l| l | l|}
				\hline
				&\multicolumn{2}{|c|}{$dataset_A$} & \multicolumn{2}{|c|}{$dataset_B$} & \multicolumn{2}{|c|}{$dataset_C$} \\
				\hline
				\textit{Method} & \textit{PICP} & \textit{MPIW}  & \textit{PICP} & \textit{MPIW} & \textit{PICP} & \textit{MPIW}\\ \hline \hline
				\textit{DAPIEN} & \textbf{94.9\%} & 0.45 & \textbf{93.5\%} & 1.77 & \textbf{97.5\%} & 18.7 \\ 
				\textit{Bootstrap} & 56.8 \% & 0.068 & 65.2\% & 0.909 & 99.9\% & 75.1 \\ \hline
			\end{tabular}
			\captionof{table}{PIs provided in Bootstrap can be too tight as in $dadaset_A$, or too wide as in $dataset_C$. DAPIEN, provides higher quality PIs because it's adaptive to the underlying noise distribution and the prior knowledge about the distribution function coded in the process.}
			\label{tab:PI_SumOfBits_2_all}
		\end{center}
	\end{table}
	
	\subsection{$dataset_A$: Sum of input bits, with added conditional white noise}
	In this experiment the synthetic generated dataset models a case where the distribution of the added noise, varies depending on the input. Initially, all unique binary vectors of length $d=10$ are generated. For each unique $x$, a vector $y$ is then generated using \eqref{dataset_1_y}. The number of elements in vector $y$ for each unique $x$ in this experiment is 20. Note that the vector $y$ may consist of equal elements, or they would vary, based on \eqref{dataset_1_err}.
	
	\begin{align}
		& x \in \{0, 1\}^d \label{dataset_1_x}\\ 
		&f(x) = x^T \cdot 1 \label{dataset_1_fx}\\ 
		&err(x) = 
		\begin{cases} \label{dataset_1_err}
			0, & \text{if} \ (f(x) \mod 2 = 0) \\
			\mathcal{N}(0, 0.2),& otherwize \\
		\end{cases}\\
		& y = f(x) + err(x) \label{dataset_1_y}
	\end{align}
	Hence the generated dataset has total $2^{10}*20 = 20480$ number of records. The dataset is then separated into 20\% test and 80\% training subsets while ensuring that there is no overlap between the subsets with regards to the values of $x$.
	
	\subsubsection{Results} \label{dataset_1_result}
\begin{figure}[ht!]
		\centering
		\subfigure[\ ]{%
			\includegraphics[width=0.490\linewidth]{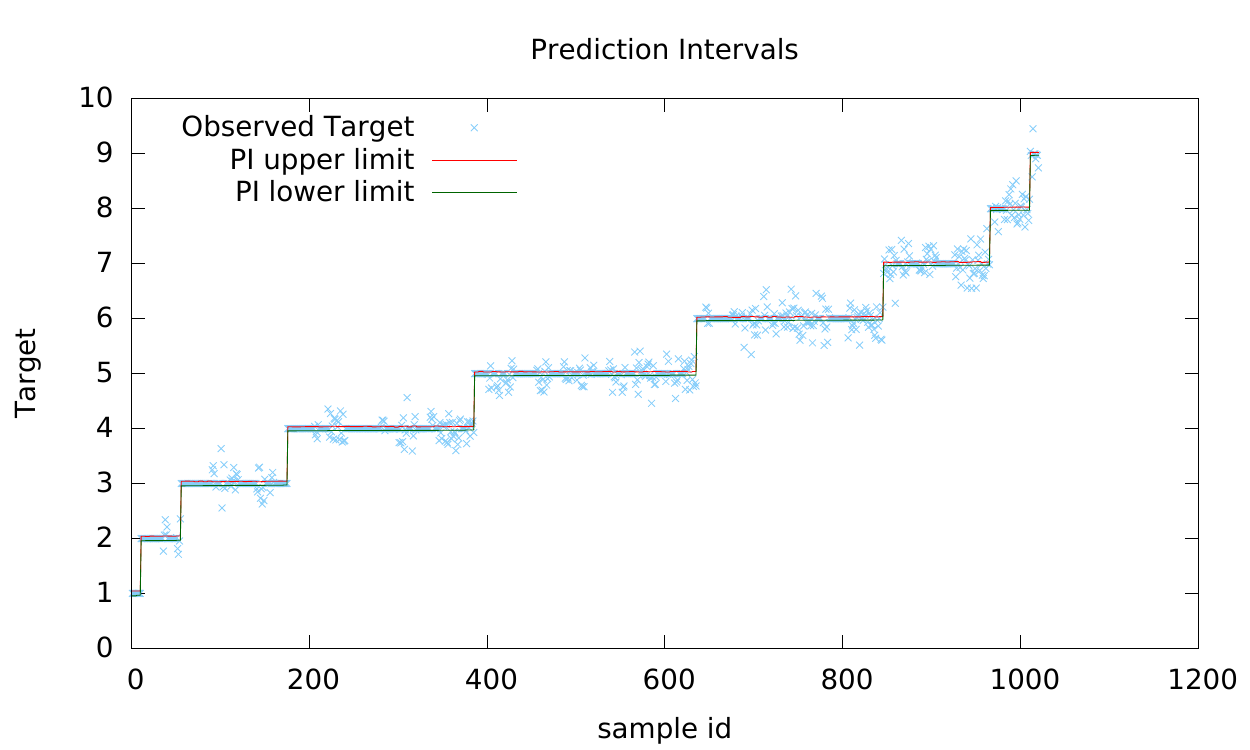}
			\label{fig:PI_SumOfBits_2_bootstrapPaired_samples}}
		\subfigure[\ ]{%
			\includegraphics[width=0.490\linewidth]{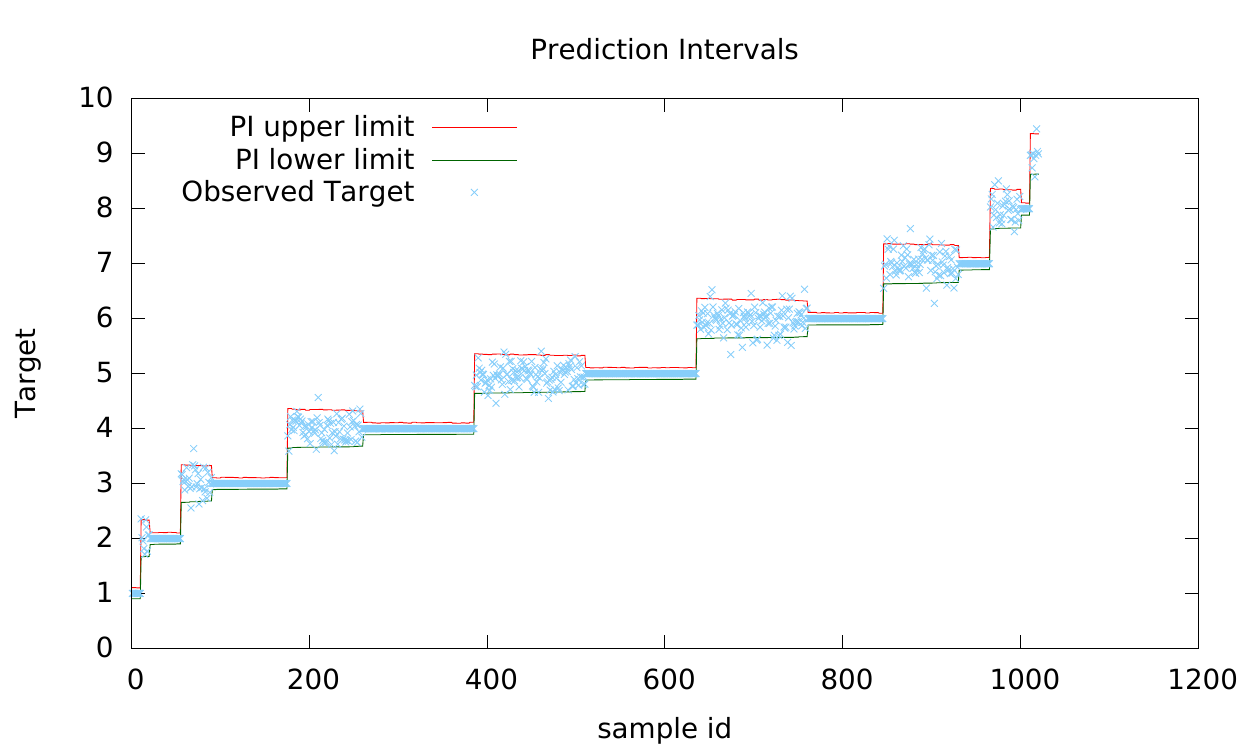}
			\label{fig:PI_SumOfBits_2_dabp1_samples}}
		
		\caption{ $dataset_A$ results: \subref{fig:PI_SumOfBits_2_bootstrapPaired_samples} shows prediction intervals for the Bootstrap method and \subref{fig:PI_SumOfBits_2_dabp1_samples} show the predicted intervals for the DAPIEN method which performs better.}
		\label{fig:PI_SumOfBits_2_all}
	\end{figure}

	Following the procedure mentioned in \ref{mpractical}, \ref{moutline}, two separate feed forward networks with no hidden layer are trained. For the FNN which predicts $\sigma$, an exponential activation function is added in order to ensure predictions are always positive. For optimizing the neural network the back-propagation \cite{hecht1989theory} is used. For finding the optimal weights, initially both Stochastic Gradient Descent (SGD) and Conjugate Gradient(CG) were tried. However for our dataset, CG with full batch, provided a much faster convergence rate. This is in accordance with optimization guidelines provided in \cite{lecun2012efficient} for small networks with small datasets. In order to avoid over-fitting, stratified 5-fold cross validation was used to select the model with lowest generalization error.
	Figure \ref{fig:PI_SumOfBits_2_all} provides qualitative comparison amongst both the Bootstrap and DAPIEN methods with respect to the provided PIs for the test data set using confidence level $95\%$.

	\graphicspath{{./figure/}}

	\subsection{$dataset_B$: Sum of input bits, with added white noise scaled by target value}
	For $dataset_B$, the records are generated similar to $dataset_A$ except for the added error as described in  \eqref{eq:dataset2_err}.
	
	\begin{equation}\label{eq:dataset2_err}
		err(x) =   f(x) \cdot \mathcal{N}(0, 0.1) 
	\end{equation} 
	\subsubsection{Results}
\begin{figure}[ht!]
		\centering
		\subfigure[\ ]{%
			\includegraphics[width=0.49\linewidth]{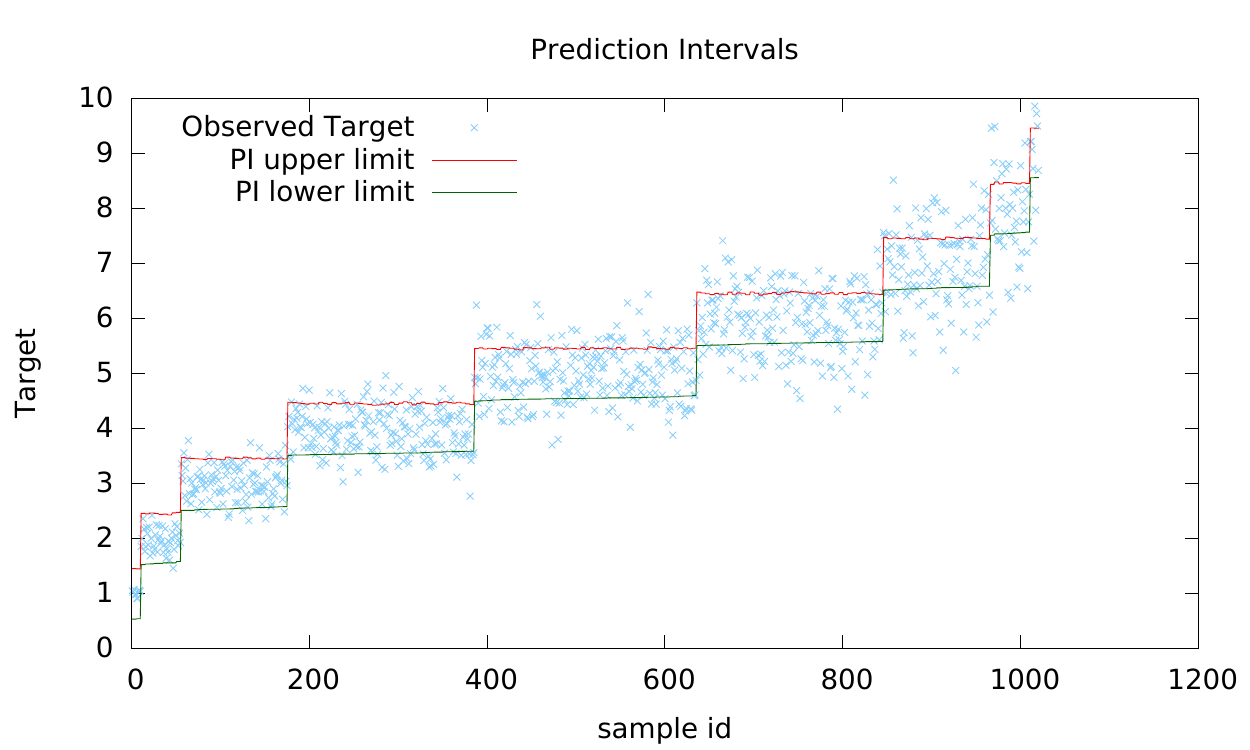}
			\label{fig:PI_SumOfBits_3_bootstrapPaired_samples}}
		\subfigure[\ ]{%
			\includegraphics[width=0.490\linewidth]{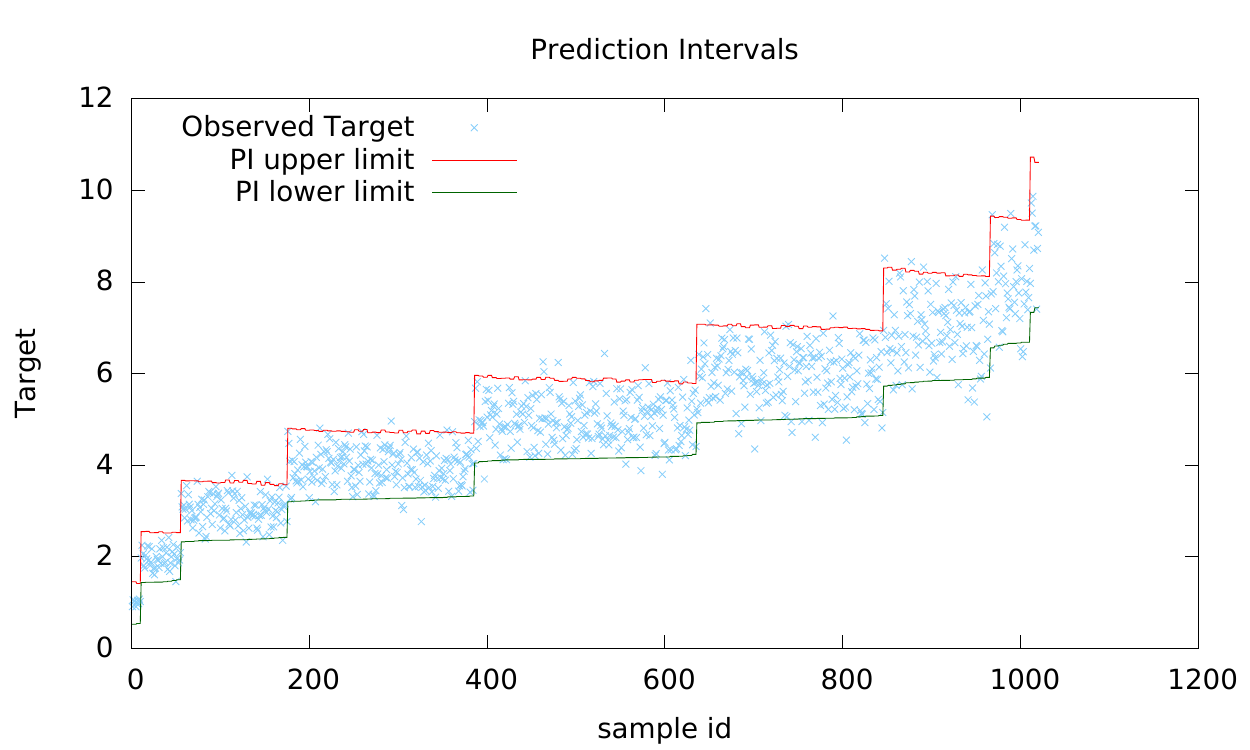}
			\label{fig:PI_SumOfBits_3_dabp1_samples}}

		\caption{ $dataset_B$ results: \subref{fig:PI_SumOfBits_3_bootstrapPaired_samples} shows prediction intervals for the Bootstrap method and \subref{fig:PI_SumOfBits_3_dabp1_samples} show the predicted intervals for the DAPIEN method which performs noticably better.}
		\label{fig:PI_SumOfBits_3_all}	
	\end{figure}

	Same FNN architecture and procedure described in \ref{dataset_1_result} is used for building the DAPIEN predictors for $dataset_B$. As the results in figure \ref{fig:PI_SumOfBits_3_all} shows, DAPIEN adjusts the PI width as the noise increases for higher target values while the Bootstrap method is providing similar PI width regardless of the target value. 
	
	\subsection{$dataset_C$: Sum of input bits, with added gamma noise scaled by target value}
	For $dataset_C$, the records are generated similar to $dataset_B$ except for the added error follows a gamma distribution as in equation \eqref{eq:dataset3_err}.
	\begin{equation}\label{eq:dataset3_err}
		err(x) =   f(x) \cdot gamma(\alpha=1, \beta=1, \mu=0) 
	\end{equation} 
	
	\subsubsection{Results}
\begin{figure}[ht!]
		\centering
		\subfigure[\ ]{%
			\includegraphics[width=0.490\linewidth]{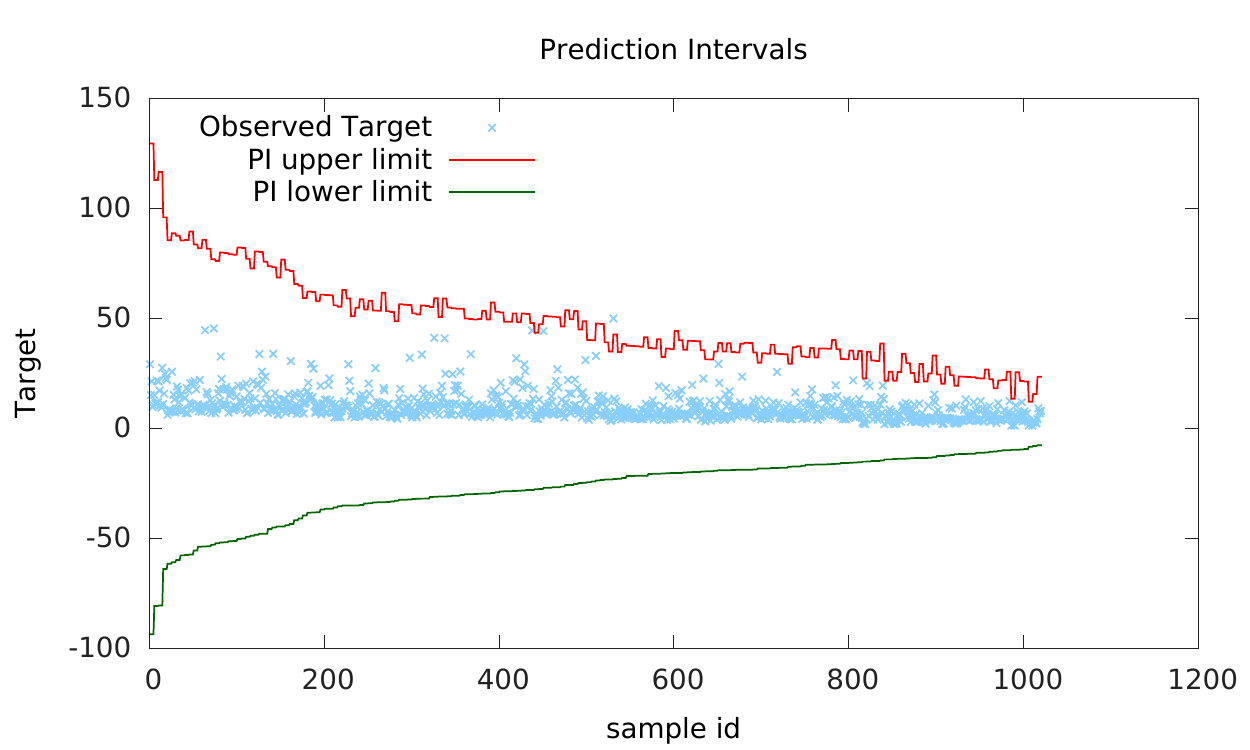}
			\label{fig:PI_SumOfBits_6_bootstrapPaired_samples}}
		\subfigure[\ ]{%
			\includegraphics[width=0.490\linewidth]{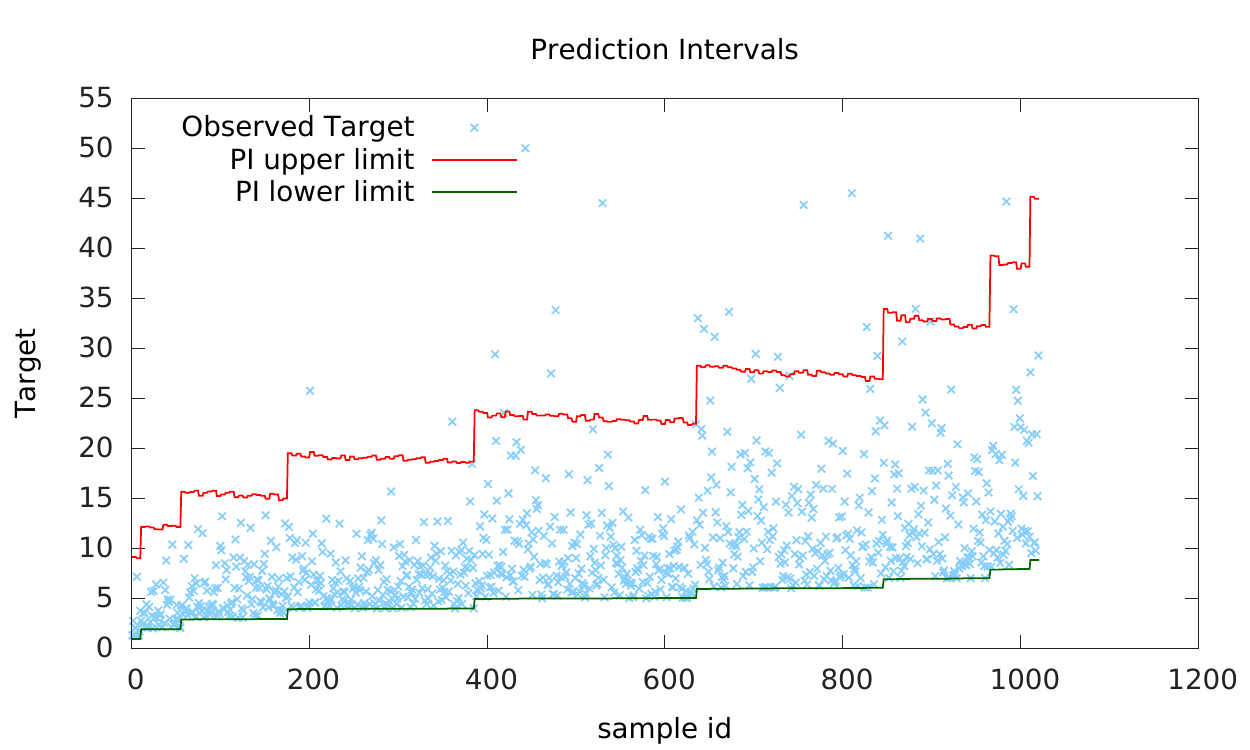}
			\label{fig:PI_SumOfBits_6_dabp1_samples}}

		\caption{ $dataset_C$ results: \subref{fig:PI_SumOfBits_6_bootstrapPaired_samples} shows prediction intervals for the Bootstrap method and \subref{fig:PI_SumOfBits_6_dabp1_samples} show the predicted intervals for the DAPIEN method which performs noticeably better.}
		\label{fig:PI_SumOfBits_6_all}	
	\end{figure}

	Following the procedure in \ref{mpracticalGamma}, three single layer FNN predictors were trained to predict the gamma distribution parameters $\alpha$, $\beta$ and $\mu$. Same optimization techniques described in \ref{dataset_1_result} were used. Figure \ref{fig:PI_SumOfBits_6_all} shows the comparison of this technique with the bootstrap method using the confidence level of 95\%. The DAPIEN method is able to recover the original function and provide appropriate prediction intervals. Although the prediction intervals provided using the bootstrap method cover 99\% of the dataset, but they are too wide.

	\section{Conclusion} \label{conclusion}
	The presented results on synthetic data, suggests that the DAPIEN, can provide accurate prediction intervals for datasets with categorical input variables. The performance of this method is particularly dependent on the appropriate selection of the distribution function. It can model a system in which, the parameters of the target distribution can change depending on the input, while the distribution function remains the same. This method can play a key role in designing more accurate predictors when input variables are solely categorical. In subsequent work, we intend to apply this technique to provide prediction intervals for gene expression levels given genetic and environmental conditions.
	\bibliographystyle{iclr2016_conference}
	\bibliography{ConfANN}
\end{document}